\documentclass{article}

\usepackage[dblblindworkshop, final]{neurips_2025}

\usepackage[utf8]{inputenc} 
\usepackage[T1]{fontenc}    
\usepackage{hyperref}       
\usepackage{url}            
\usepackage{booktabs}       
\usepackage{amsfonts}       
\usepackage{nicefrac}       
\usepackage{microtype}      
\usepackage{natbib}

\usepackage{amsmath}
\usepackage{float}
\usepackage{graphicx}
\usepackage{algorithm}
\usepackage{algpseudocode}
\usepackage{xspace}
\usepackage{stmaryrd}
\usepackage{enumitem}
    \setlist[enumerate, 1]
        {%
            leftmargin = 5.0mm, 
            labelsep = * 
        }

\usepackage[dvipsnames]{xcolor}
\newcommand{\ABCPearson}{\textbf{\color{red}ABC-Pearson}\xspace}
\newcommand{\ABCCNN}{\textbf{\color{blue}ABC-CNN}\xspace}
\newcommand{\SBICNN}{\textbf{\color{ForestGreen}SBI-CNN}\xspace}

\title{Simulation-based inference of yeast centromeres}

\author{Eloïse Touron \\
        Univ. Grenoble Alpes, Inria \\
CNRS, Grenoble INP, LJK, France \\
        \texttt{eloise.touron@inria.fr} \\
      \And Pedro L. C. Rodrigues\\
        Univ. Grenoble Alpes, Inria \\
CNRS, Grenoble INP, LJK, France \\
        \texttt{pedro.rodrigues@inria.fr} \\
    \And Julyan Arbel\\
        Univ. Grenoble Alpes, Inria \\
CNRS, Grenoble INP, LJK, France \\
        \texttt{julyan.arbel@inria.fr} \\
    \And Nelle Varoquaux\\
    TIMC, Univ. Grenoble Alpes\\
    CNRS, Grenoble INP, France \\
    \texttt{nelle.varoquaux@univ-grenoble-alpes.fr}
    \And Michael Arbel\\
        Univ. Grenoble Alpes, Inria \\
CNRS, Grenoble INP, LJK, France \\
        \texttt{michael.arbel@inria.fr} \\
}

\begin{document}

\maketitle

\begin{abstract}
  The chromatin folding and the spatial arrangement of chromosomes in the cell play a crucial role in DNA replication and genes expression. An improper chromatin folding could lead to malfunctions and, over time, diseases. For eukaryotes,  centromeres are essential for proper chromosome segregation and folding. Despite extensive research using \textit{de novo} sequencing of genomes and annotation analysis, centromere locations in yeasts remain difficult to infer and are still unknown in most species. Recently, genome-wide chromosome conformation capture coupled with next-generation sequencing (Hi-C) has become one of the leading methods to investigate chromosome structures. Some recent studies have used Hi-C data to give a point estimate of each centromere, but those approaches highly rely on a good pre-localization. Here, we present a novel approach that infers in a {stochastic manner} the locations of all centromeres in budding yeast based on both the experimental Hi-C map and {simulated contact maps}. 
\end{abstract}

\section{Introduction}

Hi-C maps have become one of the main assets in understanding DNA folding, notably through the study of chromatin loops or topologically associated domains (TADs) in mammalian cells \cite{tads, loops}. These maps capture the contact counts between fragments of chromosomes among a population of DNA into a 2D squared and symmetric matrix made of cis- and trans- blocks of interactions. 

Beside chromatin loops or TADs, centromeres are also of great interest to genome structure investigation due to their essential role in many biological processes: they facilitate chromosome segregation through the formation of the kinetochore~\cite{kinetochore} during mitosis and meiosis, and act as key regulators of genome stability via the prevention of chromosome breakage. In yeasts, centromeres are highly compact regions spanning about 125 base pairs (bp) \cite{centrosize} and tend to cluster near the spindle pole body within the nucleus. This clustering results in a distinct peak in the trans-contact counts Hi-C matrices, centered at the position of each centromere pair. Many studies have attempted to annotate yeast centromeres, usually through Fluorescent In Situ Hybridization (FISH) \cite{FISH} or chromatin immunoprecipitation (ChIP) \cite{CHIP}. However, these approaches often remain imprecise and may even fail to infer centromeres for some species \cite{inferfail}. To bypass these limitations, newer methods have been proposed, which use Hi-C contact maps \cite{nelly, nelle}. Working directly with a Hi-C contact map or with the corresponding Pearson correlation matrix, they fit a Gaussian to each interaction peak to precisely infer centromere locations. Beyond the optimization of a non-convex function, these methods highly rely on good pre-localization of the centromeres to be precise and output only a point estimate of each centromere whereas it is actually a whole segment of chromosome.

We propose a novel approach to infer centromere positions that differs from existing methods in two key aspects. Firstly, we adopt a {stochastic approach} by quantifying the uncertainties about the centromere candidates we infer. Secondly, the inference processes are not only based on a reference Hi-C matrix (denoted $C_{\text{ref}}$) but also on {simulated contacts maps} (denoted $C$). We adopt a {Bayesian} approach where centromere positions (denoted $\theta$) are sampled from a prior distribution and the contact maps $C$ are generated from a {custom-designed simplified simulator}. We estimate the posterior distribution $p(\theta|C_\text{ref})$, which amounts to solving the following inverse problem: \begin{center}
\small
\boxed{
    \textbf{Given contact map $C_\mathrm{ref}$, what are the most probable centromere positions $\theta$ to have generated it?} 
}
\end{center}

\section{Methods}
\subsection{Framework}
\paragraph{Setting.} We work with the budding yeast \textit{Saccharomyces cerevisiae} for which the positions of its 16 centromeres and length of each chromosome (in base pairs) are known. We used data from Duan et al. \cite{duan} to construct the reference Hi-C contact map. Centromere candidates $\theta = (\theta_1,\ldots, \theta_{16})$ are sampled from a prior distribution $p(\theta)$ that is poorly informative, e.g. a multivariate uniform distribution where the interval is the length of each chromosome in each dimension. To simulate contact maps, we designed a simplified simulator (described below) that takes $\theta$ as input and directly outputs realistic $C$ without simulating any DNA folding. Other studies have used simulators to do Bayesian inference of biological elements \cite{arbona}, but those biological simulators try to simulate the 3D folding of the chromatin in the nucleus before computing the corresponding contact map. This renders them too slow and unnecessarily complex if we want to have many contact maps in a reasonable time.

\paragraph{Contact maps and data normalization.}
The contact map $C$ projects the information contained in a population of 3D chromatin foldings into a 2D squared and {symmetric} matrix made of {cis-} (or intra-chromosomal) and {trans-} (or inter-chromosomal) blocks of interactions between pairs of chromosomes. To construct it, we cut each chromosome into genomic windows of a given length (called resolution), e.g. 32 kilobases (kb). Each matrix entry is then a non-negative number, called the contact count, representing the number of times a given window was in contact with another one over the population (see Appendix \ref{sec:hic_appendix} and Figure \ref{fig:contactmap_schema} for more details). 

During inference, we use a reference Hi-C map $C_{\text{ref}}$ and simulate synthetic contact maps $C$. Hi-C contact maps have many biases due to sequencing and mapping errors or to the inherent structure of the chromatin \cite{hicnorm}. Therefore, $C_{\text{ref}}$ is actually a normalized Hi-C map, where the normalization corrects those biases, iteratively forcing all rows and columns to sum up to one \cite{hicnorm}. The quality of the contact map depends on the chosen resolution and the signal-to-noise ratio gets smaller if we work at higher resolution. We thus choose to set the contact maps at resolution 32 kb. In yeasts, the main informative part about centromeres rely on the upper trans-contact blocks: the matrix of contacts between chromosomes $i$ and $j$ contains an enrichment of interactions at the location of both centromeres $(\theta_i, \theta_j)$. 

\paragraph{Simulator.}
\label{sec:simulator}
We exploit the structure of yeast contact maps to design a very efficient simulator that directly {creates the upper trans-contact blocks} given its centromeres positions $\theta$.
At the centromere positions, the chromatin has a brush-like organization: chromosomal regions near the centromeres often enter in contact over the population whereas the further we move away from the centromeres, the rarer the contacts become. To mimic this effect, we simulate a Gaussian spot at the position ($\theta_i, \theta_j$) for each trans-contact block.
Between chromosomes, we also observe rare interactions over the population that we reproduce by adding Gaussian noise to all the trans-contacts blocks up to $10\%$ of the maximal contact count (see Appendix~\ref{sec:simulator_appendix} with Algorithm~\ref{alg:simu} and Figure~\ref{fig:C_simu} for more details).

\subsection{Simulation-based inference}

Our goal is to infer $\theta$ from $C_\text{ref}$ using a probabilistic framework based on simulations. The usual way for doing so would be to search for the most appropriate $\theta$ for a given $C_\text{ref}$ by maximizing the likelihood: 
\begin{equation*}
    \hat{\theta} = \underset{\theta \in \Omega}{\text{argmax}}\ \log p(C_\text{ref}|\theta)~.
\end{equation*} 
However, as the simulator is often very complex (e.g. biological simulators that try to mimic a 3D folding of the DNA given a set of constraints), the {likelihood} $p(C|\theta)$ may be {intractable}. As such, we directly {target the posterior density} $p(\theta|C_\text{ref})$ using data from the joint model ($\theta, C) \sim p(\theta)p(C | \theta)$, either via {approximate Bayesian computation} (Sequential Monte-Carlo ABC: SMC-ABC) or by estimating the posterior density with a conditional normalizing flow (Sequential neural posterior estimation: SNPE) \cite{snpea, snpec} .

\paragraph{SMC-ABC.}
We use a variant of ABC coupled with sequential Monte-Carlo (SMC)~\cite{smcabc}. It consists of multiple rounds of ABC where, at each round, relevant $\{\theta^{k,*}\}_k$ are selected from the training set $\{(\theta^n, C^n)\}_n$ depending on a closeness criterion between $C$ and $C_\text{ref}$. We then associate weights $\{w^k\}_k$ to those selected $\{\theta^{k,*}\}_k$, and use the set $\{(\theta^{k,*}, w^k)\}_k$ to create the next population of $\{\theta^n\}_n$ for the next round of ABC. This sequential approach enables us to refine the relevant $\theta$ at each round.
However, we need to {define a metric} for discriminating $(\theta^n, \theta^m)$ based on their associated observations $(C^n, C^m)$.

\begin{enumerate}
\item[]\textbf{\underline{Metric 1} : Pearson correlation -- \ABCPearson.}
To measure the closeness between $C$ and $C_\text{ref}$, the Pearson correlation is commonly used~\cite{tjong, nelly, nelle}. We find that the {vector-based Pearson correlation} averaged over all trans-contacts blocks is the most discriminative metric: each trans-contacts block of $C$ and $C_\text{ref}$ is vectorized and the Pearson correlation is computed between both. We then average all the correlations over the trans-contacts blocks (see Algorithm ~\ref{alg:smcabc} in Appendix \ref{sec:SMCABCPearson}). However, this metric is fine-tuned to this specific inference task.

\item[]\textbf{\underline{Metric 2} : Data-driven summary statistic -- \ABCCNN.}
Instead of looking for a specific metric to compare $C$ to $C_\text{ref}$, we choose to use the classical $l^2$-norm. For this, we need a summary statistic $S$ that will extract the main features of $C$ and project it into a low-dimensional vector. One relevant candidate for a summary statistic is $\mathbb{E}\left[\theta|C\right]$ because with this one: 
\begin{equation*}
   \mathbb{E}\left[\theta \mid \Vert S(C)-S(C_\text{ref})\Vert \leq \epsilon\right] \underset{\epsilon \to 0}{\to} \mathbb{E}\left[\theta|C_\text{ref}\right]~,
\end{equation*}where $\epsilon$ is the ABC-threshold.\\
When $\epsilon \to 0$, we don't lose any first-order information when summarizing $C$ \cite{summstatabc}. Moreover, $\mathbb{E}\left[\theta|C\right]$ is the analytical solution of the regression of $\theta$ on $C$ i.e. \begin{equation}
    \mathbb{E}\left[\theta|C\right] = \underset{S \in \mathcal{F}}{\text{argmin}}\ \mathbb{E}\left[ \Vert S(C)-\theta \Vert ^2_2\right]~,
\end{equation} where $\mathcal{F}$ is the set of square integrable functions. As this statistic is unavailable, we learn it via a (deep) neural network (DNN) $S_{\phi}$ with parameters $\phi$ \cite{summstatabc}. The DNN encoding $S_{\phi}$ is composed of a {convolutional neural network (CNN)} followed by a {multi-layer perceptron (MLP)}. Using Monte Carlo estimator of (1) with $N$ samples $(\theta^n, C^n) \sim p(\theta)p(C|\theta)$, the DNN loss to be minimized in $\phi$ is then \begin{equation*}
    \widehat{\mathcal{L}}_{\text{DNN}}(\phi)=\frac 1N \sum_{1\leq n\leq N}  \Vert S_{\phi}(C^n) - \theta^n \Vert ^2_ 2~.
\end{equation*}
For large $N$, we expect $S_{\phi}(C) \approx \mathbb{E}\left[ \theta|C \right]$. This approach has two phases: first learn $S_\phi$ by minimizing $\widehat{\mathcal{L}}_{\text{DNN}}(\phi)$, then run sequential ABC with this summary statistic and the $l^2$-norm as discriminative criterion (see Algorithm~\ref{alg:abc-cnn} in Appendix~\ref{sec:ABCCNN_appendix}).

\end{enumerate}

\paragraph{SNPE -- \SBICNN.}

SMC-ABC yields only {samples from the target posterior distribution} $p(\theta|C_\text{ref})$, but evaluating log-probabilities can be useful for downstream tasks. In contrast, a conditional normalizing flow $p_{\psi}(. | .)$~\cite{snpea, snpec} used to estimate the posterior distribution can both easily sample from the posterior and return the values of its log-probabilities. To ensure that $p_{\psi}(\theta|C_\text{ref})$ is close to $p(\theta | C_\text{ref})$, we minimize their Kullback--Leibler divergence ($D_\mathrm{KL}$) averaged over the observations $C$ as per 
\begin{equation*}
    \mathbb{E}_C \big[ D_\mathrm{KL}\big( p(\cdot | C) \Vert p_{\psi}(\cdot|C) \big) \big].
\end{equation*}
After simplifications and using a Monte Carlo estimator, the flow is trained to minimize
\begin{equation*}
    \widehat{\mathcal{L}}_{\text{NPE}}(\psi)=-\frac 1 N \sum_n \log(p_{\psi}(\theta^n|C^n)) \text{ , } (\theta^n, C^n) \sim p(\theta)p(C|\theta).
\end{equation*}
Once trained, we obtain an amortized estimator of the posterior densities $p(\theta | C)$ valid for any $C$. We just have to plug in $C_\text{ref}$ to get the estimated posterior density $p_{\psi}(.|C_\text{ref})$ (see Algorithm~\ref{snpe} in Appendix~\ref{sec:snpe_appendix}). Since we are actually interested in the posterior at $C_\text{ref}$,  
parameters $\theta$ with very low posterior density may not be useful for learning $\psi$. Thus, we consider a sequential approach with several rounds of NPE to get an iterative refinement of the posterior estimate~\cite{snpec}. From the second round, $\theta^n$ are sampled from the latest estimated posterior found instead of the prior. This way, training samples are more informative about $C_\text{ref}$, gradually improving the learning of $\psi$. When the observations $C$ are high-dimensional (e.g. 2D-matrices), we encode them in a summary statistic $S$ using a convolutional neural network.

\section{Numerical experiments}
We showcase our methodology on two settings involving the genome of the yeast \textit{S. cerevisiae}, for which we have access to the true position of all its centromeres: firstly, we run our inference pipeline on only \textit{S. cerevisiae}'s first three chromosomes, secondly on its whole genome (16 chromosomes). We assess the performance of different inference methods by comparing their approximate posterior distributions to a ground-truth distribution consisting of Diracs on each dimension located at the true centromere positions. All experiments can be run on the CPU of a laptop, requiring $\sim 1$~h for the small genome and $\sim 5$~h for the entire genome.

\subsection{Study case -- small genome (3 chromosomes)}
In low-dimensional settings, we can jointly infer $\theta$ given the entire contact map $C_\text{ref}$. For each inference method, we consider 11 rounds, each with a training dataset $\{(\theta^n, C^n)\}_n$ of size $10^3$. The summary statistic $S_\phi$ is pre-trained using a different training set of size $5\times10^3$ with the optimizer \texttt{Adam} and a fixed learning rate $5\times 10^{-4}$. 
For SNPE, we use a masked autoregressive flow (MAF) and the version SNPE-C~\cite{snpec} from the Python package \texttt{sbi}~\cite{sbipackage2024} (see Appendix \ref{sec:3_chr_appendix} for more details). 
 

\begin{figure}[t]
    \begin{minipage}{0.68\linewidth}
    \centering    
    \includegraphics[width=\columnwidth]{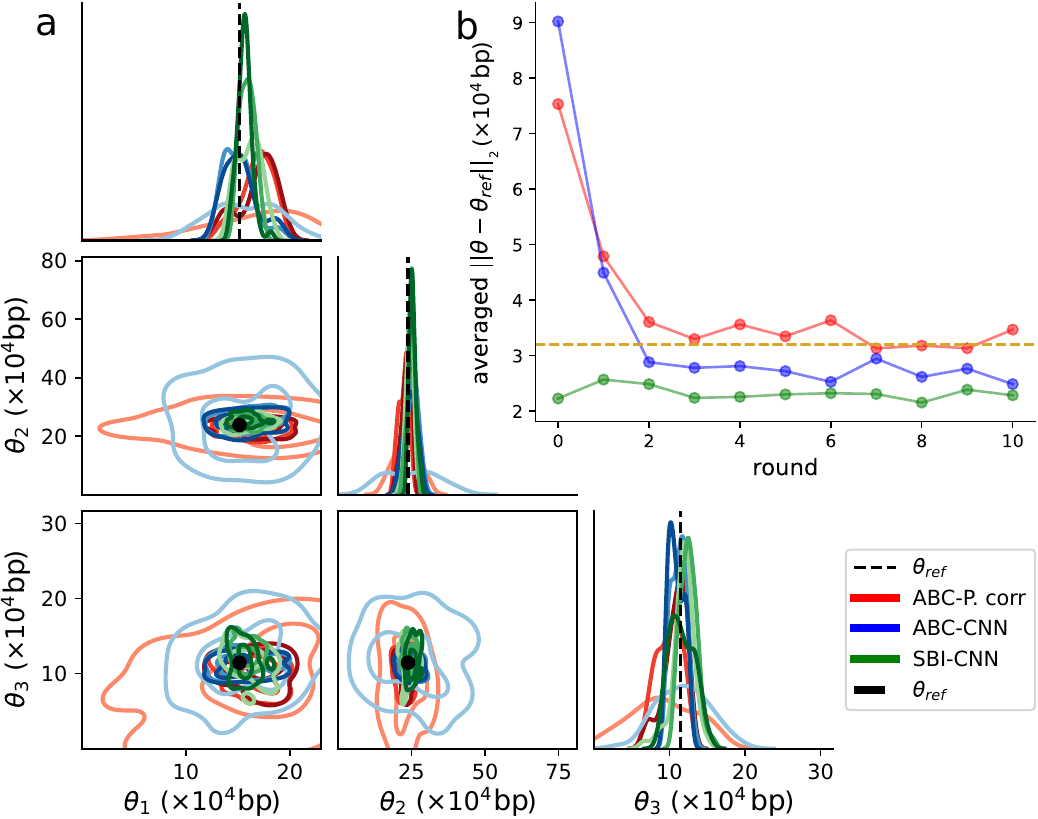}
    \end{minipage}
    \hfill
    \begin{minipage}{0.30\linewidth}
    \caption{\label{fig:toy_model_3_chr}{\small Inference using \ABCPearson, \ABCCNN, and \SBICNN  (\textbf{a}). Color shades increase from lightest to darkest across rounds. Densities are estimated with the $5\%$ best $\theta$ according to the ABC criterion or sampled from the flow. 
    We also report the mean Euclidean distance between $\theta$ and $\theta_\text{ref}$, computed over the $5\%$ best-performing samples in the top right corner (\textbf{b}). The horizontal dashed line stands for the resolution of the contact map $C_\text{ref}$ (in bp) in the top right figure. Results with \SBICNN are uniformly better and both approaches based on data-driven summary statistics have errors smaller than the resolution of the contact maps.
    }}
    \end{minipage}
    \vspace{-1em}
\end{figure}

\subsection{High dimensional problem -- whole genome (16 chromosomes)}
When extending the analysis to the entire genome, we end up facing the curse of dimensionality: the space of parameters $\theta$ becomes too large to cover with few simulations and the contact maps $C$ are too big. As such, the resulting neural network encoding $S_\phi$ has too many parameters to be optimized. To reduce the dimension of the problem, we run $16$ {parallel inferences} (one per dimension of $\theta$) extracting each time only the informative part of the contact maps. With this approach, the space of $\theta$ is cut into several chromosome-length 1D intervals reducing the train set size. 
Let $C_i$ and $C_{\text{ref},i}$ be the $i^\mathrm{th}$ lines of blocks of matrices $C$ and $C_{\text{ref}}$, respectively. 
To infer $\theta_i$ with \ABCPearson, we compute the vector-based Pearson correlation averaged over all blocks between $C_i$ and $C_{\text{ref},i}$. Concerning the data-driven summary statistic approaches \ABCCNN and \SBICNN, the summary statistic $S_{\phi_i}$ also tries to project $C_i$ to $\theta_i$. 

Using the redundancy of the data between rows of blocks, and to minimize the number of parameters of $\{S_{\phi_i}\}_i$, we consider a shared architecture where the CNN parameters are shared between chromosomes and the MLP ones are chromosome-specific. For \SBICNN, we learn $16$ normalizing flows $p_{\psi_i}(\theta_i | S_{\phi_i}(C_{\text{ref}, i}))$ (see Appendix \ref{sec:16_chr_inference} for more details and results). 

\begin{figure}[t]
    \begin{minipage}{0.68\linewidth}
    \centering    
    \includegraphics[width=\columnwidth]{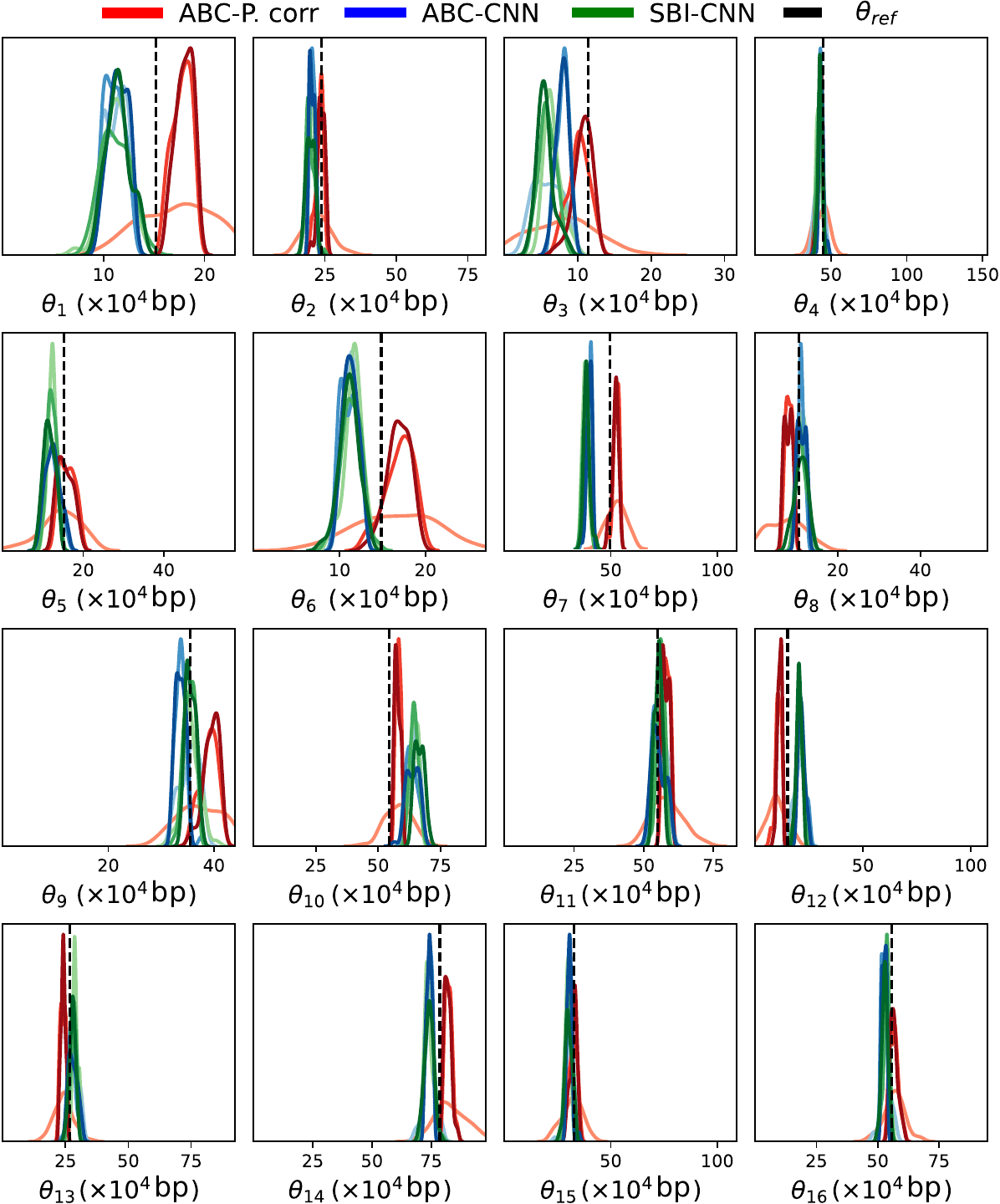}
    \end{minipage}
    \hfill
    \begin{minipage}{0.30\linewidth}
    \caption{\label{fig:model_16_chr}{\small Inference using \ABCPearson, \ABCCNN, and \SBICNN. Color shades increase from lightest to darkest across rounds. Densities are estimated with the $5\%$ best $\theta$ according to the ABC criterion or sampled from the flow. 
    In some dimensions, the densities are very peaky and centered around $\theta_i$ (e.g. chromosome 4, 13, 15) but in others, the inference is not precise (e.g. chromosome 1, 6, 10). Data-driven summary statistics approaches do not outperform Pearson correlation-based method.
    }}
    \end{minipage}
    \vspace{-1em}
\end{figure}

\section{Discussion}
We present a novel methodology to infer the positions of the centromeres of the yeast \textit{S. cerevisiae} using Hi-C contact maps. The probabilistic framework that we use allows us to quantify the uncertainty about the centromere candidates. Our entire inference pipeline is based on a large number of simulations relating centromere positions and contact maps. To mitigate computing bottlenecks, we have designed a {simplified} but efficient {simulator} that yields very convincing results when coupled with {inferences on real experimental data}.

In the case of a small genome, we obtained accurate inference of the centromere positions (Figure~\ref{fig:toy_model_3_chr}). The estimated densities for the summary statistic-based methods (\ABCCNN and \SBICNN) are not very biased and peaky around the ground truth. In each dimension, $\theta$ is estimated at a precision under the resolution of the contact map $C_\text{ref}$ (Figure~\ref{fig:metrics_curves}a) and the Euclidean distance to $\theta_\text{ref}$ is also under the resolution (Figure~\ref{fig:toy_model_3_chr}b). Moreover, \SBICNN outperforms \ABCCNN that itself outperforms \ABCPearson, reinforcing the use of a summary statistic and the flexibility of the normalizing flows. In the case of the whole genome, our approaches are not as accurate and could be improved. In some dimensions, $\theta$ is estimated at a precision under the resolution (Figure~\ref{fig:metrics_16_chr}b), and we obtain peaky densities but in others the inference is not precise (Figure~\ref{fig:model_16_chr}).

An advantage of our method is that we do not rely on any initialization or pre-localization: instead, we use an uninformative prior, setting each centromere randomly in the range of its corresponding chromosome. Also, our approach is naturally scalable: the pre-trained summary statistic could be reused for inference on centromeres of others yeasts without any re-training. To improve our approach, we will focus our efforts in developing a summary statistic independent of the size of the genome via notably the use of transformer architectures~\cite{Vaswani2017}. 

\begin{ack}
    This work was supported by the ANR project BONSAI (grant ANR-23-CE23-0012-01) and by the ANR project Bayes-Duality (grant ANR-21-JSTM-0001).
\end{ack}

\bibliographystyle{plain} 
\bibliography{biblio}


\newpage
\appendix

\section{Contact maps \label{sec:hic_appendix}}
A contact map summarizes all the chromatin contacts observed over a population of DNA configurations. To construct it, we define the resolution of the map (the length of the chromosome fragment that will represent one pixel in the map). Each chromosome is then cut into fragments and each entry of the map represents the contact counts of any fragment with another over the population of DNA. This creates a matrix by blocks of interactions between chromosomes. Usually, we represent them by a heatmap.
\begin{figure}[H]
    \centering
    \includegraphics[width=9cm, height=3.5cm]{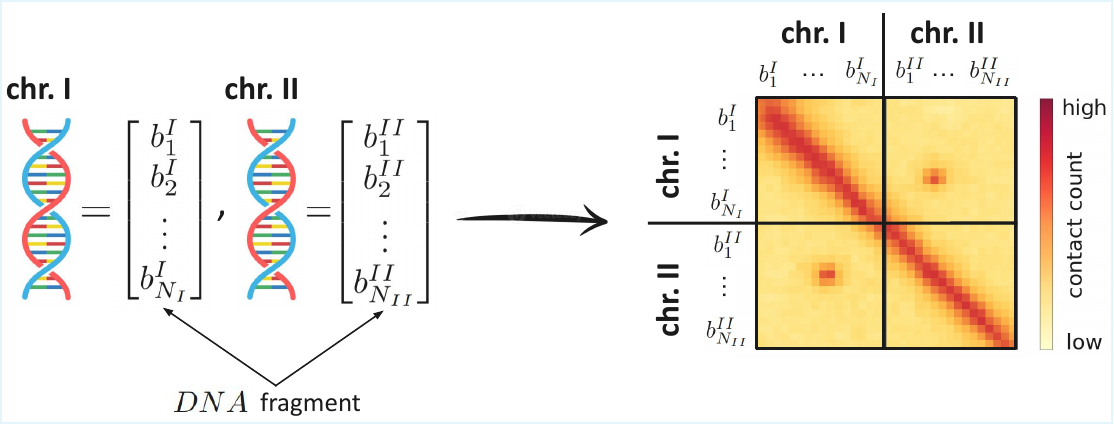}
    \caption{\small Process to construct a contact map in the case of $2$ chromosomes. \label{fig:contactmap_schema}}
\end{figure}

\section{The simulator \label{sec:simulator_appendix}}
The goal of the simulator is to create the upper trans-contact blocks of a contact map $C$ rapidly given the centromere positions $\theta$. We want to mimic the peak of interaction that appears in those blocks, as well as some rare interactions that can occur among the population of DNA.\\
Given the $L$ chromosome lengths in bp $\{l_i \}_{1 \leq i \leq L}$, the centromere positions $\theta$ are sampled from the prior $\mathcal{U}(\underset{1 \leq i \leq L}{\prod} [1, l_i-1])$. To create each contact map $C$, the process is described in Algorithm~\ref{alg:simu}. 

\begin{algorithm}[H]
\caption{Simulator of contact maps \label{alg:simu}}
\begin{algorithmic}
\State \textbf{Input}: $L$ chromosome lengths in bp $\{l_i \}_{1 \leq i \leq L}$, resolution of the contact map in bp $r$ (e.g. $r = 32$ kb), centromere positions $\theta$
\State \textbf{Return}: the upper trans-contact blocks of a simulated contact map $C$ at the resolution $r$ bp.
\State
\State choose the size of the peaks of interaction: sample $\sigma^2$ from $\mathcal{U}(0.1,10)$
\State choose the intensity of interaction $\alpha$ to simulate the DNA population size: sample $\alpha$ from $\mathcal{U}(\llbracket 1, 1000\rrbracket)$
\For{each chromosome pair $(i,j)$}
\State define a block of interaction $C_{ij}$ of size ($\frac{l_i}{r}, \frac{l_j}{r}$)
\State define the center of the peak $(\theta_i, \theta_j)$ 
\State apply Gaussian density $\mathcal{N}( (\theta_i/r, \theta_j/r) , \sigma^2 )$ to the pixels of the block $C_{ij}$ 
\State multiply each pixel of $C_{ij}$  by the intensity factor $\alpha$
\State add Gaussian noise up to $10\%$ of the maximal value of $C_{ij}$ to mimic the rare contacts:
\State construct a random matrix $M_{ij}$ of size ($\frac{l_i}{r}, \frac{l_j}{r}$) where each pixel is sampled from 
\State $\mathcal{N}(\max(C_{ij}) \times 0.05, (\max(C_{ij}) \times 0.05)^2)$, then add $M_{ij}$ to $C_{ij}$
\EndFor
\State \textbf{return} a simulated contact map $C$ at resolution $r$ bp
\end{algorithmic}
\end{algorithm}

\begin{figure}[H]
    \centering
    \includegraphics[width=14cm, height=9cm]{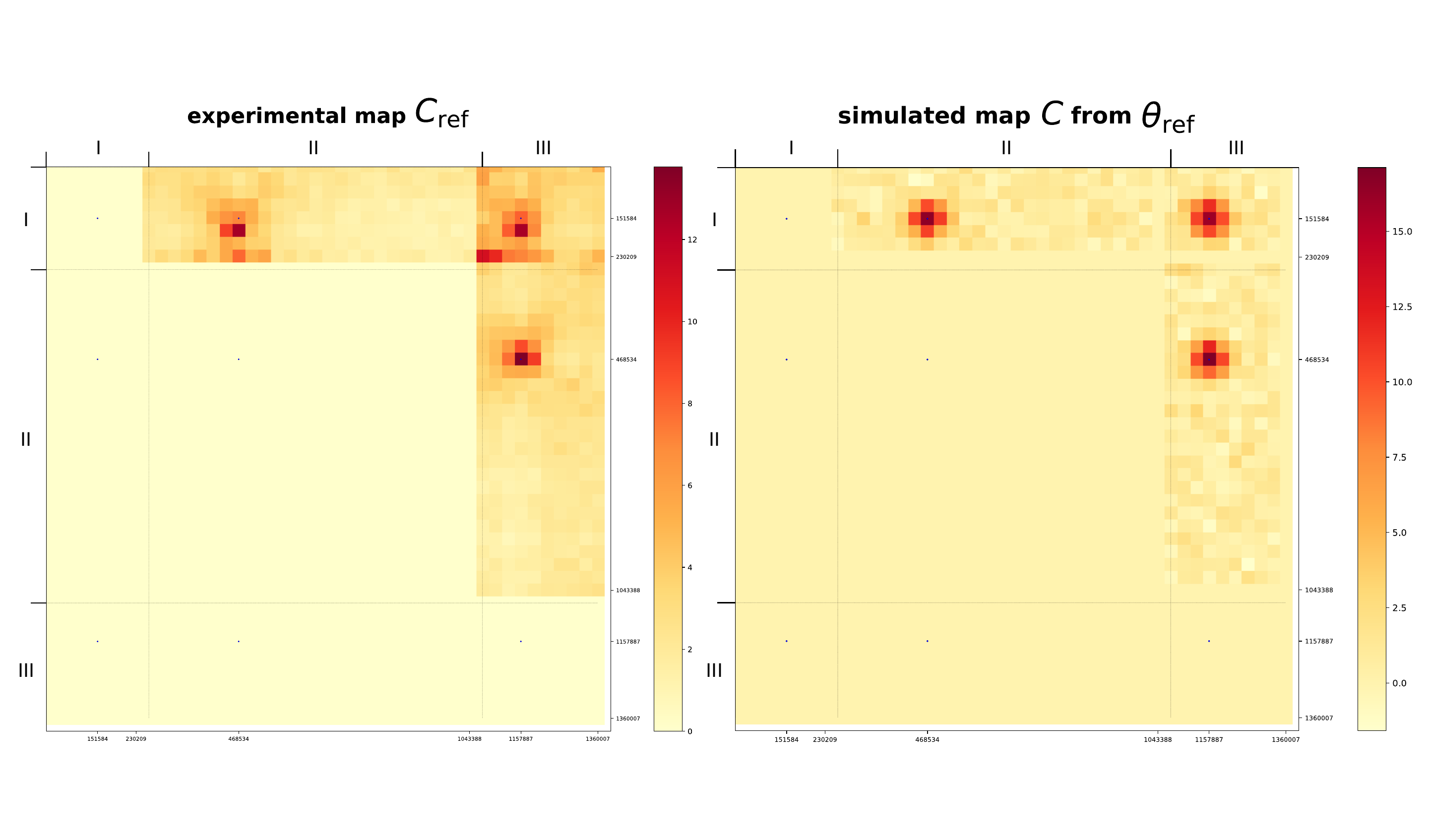}
    \caption{\small Hi-C map and our simulated map in the case of a small genome (resolution 32 kb). \label{fig:C_simu}}
\end{figure}
\begin{figure}[H]
    \begin{minipage}{0.68\linewidth}
    \centering    
    \includegraphics[width=\columnwidth]{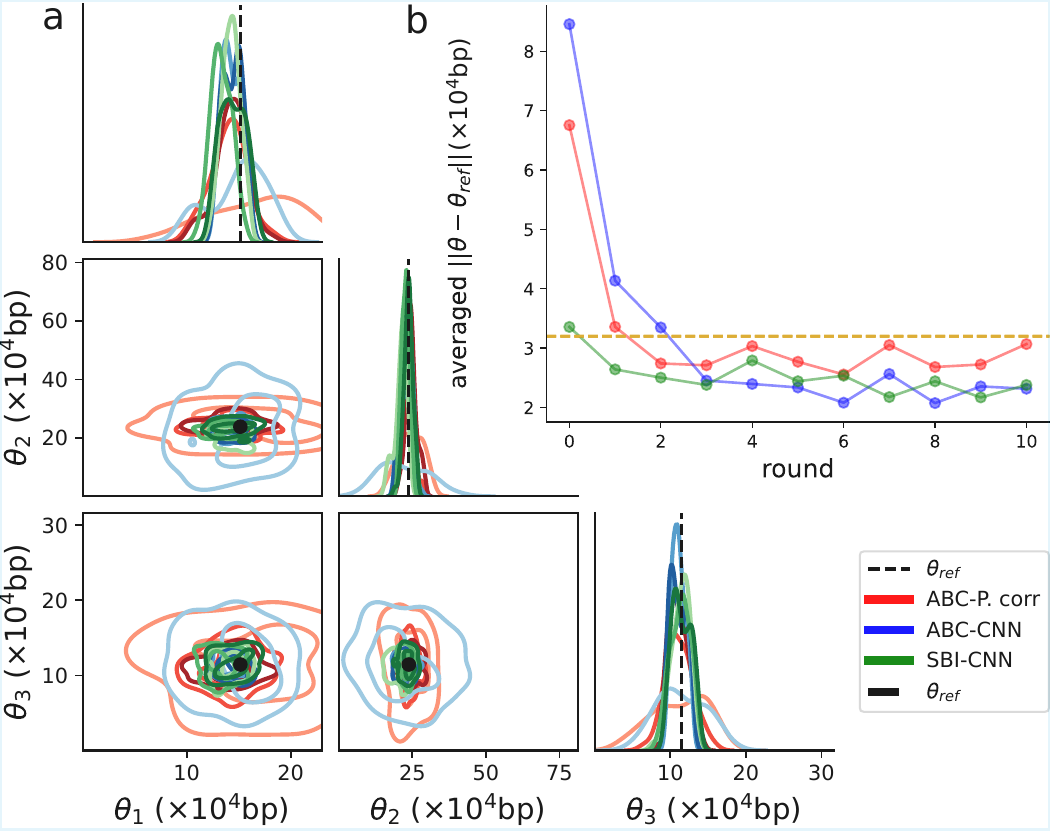}
    \end{minipage}
    \hfill
    \begin{minipage}{0.30\linewidth}
    \caption{\label{fig:toy_model_3_chr_synthetic}{\small Inference using \ABCPearson, \ABCCNN, and \SBICNN from \underline{synthetic data} (\textbf{a}). Color shades increase from lightest to darkest across rounds. Densities are estimated with the $5\%$ best $\theta$ according to the ABC criterion or sampled from the flow. 
    We also report the mean Euclidean distance between $\theta$ and $\theta_\text{ref}$, computed over the $5\%$ best-performing samples in the top right corner (\textbf{b}). The horizontal dashed line stands for the resolution of the contact map $C$ (in bp) in the top right figure. Results with data-driven summary statistics approaches are uniformly better even if all approaches have errors smaller than the resolution of the contact maps.
    }}
    \end{minipage}
    \vspace{-1em}
\end{figure}

Our simulator outputs contacts maps that present some dissimilarities with Hi-C maps. If we compute the row-based averaged Pearson correlation between $C_\text{ref}$ and $C$ simulated from $\theta_\text{ref}$ as in \cite{tjong}, we get a correlation of $0.18$ in the case of $3$ chromosomes and $0.12$ in the case of $16$ chromosomes, which is quite low.\\
However, concerning the inference task, our simulator estimates the centromere positions $\theta$ nearly as well on synthetic data (Figure~\ref{fig:toy_model_3_chr_synthetic}) as on Hi-C data (Figure~\ref{fig:toy_model_3_chr}).  

\newpage
\section{SMC-ABC}
\subsection{With the metric Pearson correlation -- \ABCPearson \label{sec:SMCABCPearson}}
One of the inference methods used is sequential ABC with the metric vector-based Pearson correlation averaged over all trans-contacts blocks. 
\begin{algorithm}[H]
\caption{SMC-ABC based on Pearson correlation inspired from \cite{smcabc} }\label{alg:smcabc}
\begin{algorithmic}
\State \textbf{Input}: $T$ rounds, prior $\pi$, train set of size $N$, acceptance size $M$, perturbation kernel $K = \mathcal{N}(., \sigma^2\text{Id})$ ($\sigma = $ resolution (bp))
\State \textbf{Return}: $\theta \sim p(\theta|\text{corr}(C,C_\text{ref}) \geq \epsilon_{\text{corr}})$
\State \textbf{round $t=0$}
\State - sample $\theta^n \sim \pi$, and $C^n \sim p(.|\theta^n) , n \in \llbracket 1, N \rrbracket$
\State - compute $\text{corr}(C^n, C_\text{ref})$ and keep the top $5\%$ of $\{\theta^n\}_n$ in terms of the highest correlation: $\{\theta^{m,0}, m \in \llbracket 1, M\rrbracket\}$
\State - compute weights $\{w^{m,0} = \frac{1}{M} , m \in \llbracket1, M\rrbracket\}$
\State \textbf{output round $t=0$}: $\{ (\theta^{m,0}, w^{m,0}) \}_{m \in \llbracket 1, M\rrbracket}$
\For{$0 < t < T$}
\State \textbf{round $t$}
\State - from the previous accepted $\{\theta^{m,t-1}\}_{m \in \llbracket 1, M \rrbracket }$, sample $\{ \bar{\theta}^k, k \in \llbracket 1, M \rrbracket\}$ from multinomial $\mathcal{M}(\{\theta^{m,t-1}\}_m, \{w^{m,t-1}\}_m)$ with replacement
\State - perturb $\frac{N}{M}$ times the $M$ samples $\bar{\theta}^k$ to have $N$ samples $\theta^n$ 
\begin{equation*}
    \theta^n \gets \bar{\theta}^k + \epsilon \text{ with } \epsilon \sim \mathcal{N}(0, \sigma^2\text{Id}) \text{ for } k=n \text{ mod } M \text{ and } n=1,...,N
\end{equation*}
\State - check that $\theta^n$ is in the prior bound otherwise, set $\theta^n \gets \bar{\theta}^k$
\State - from this set $\{\theta^n\}_{n \in \llbracket 1,N\rrbracket}$, sample $C^n \sim p(.|\theta^n) , n \in \llbracket1, N\rrbracket$
\State - compute $\text{corr}(C^n, C_\text{ref})$ and keep the top $5\%$ of $\{\theta^n\}_n$ in terms of the highest correlation: \{$\theta^{m,t}, m \in \llbracket1, M\rrbracket\}$
\State - compute corresponding weights 
\begin{equation*}
     w^{m,t} = \frac{\pi(\theta^{m,t})}{\sum_{k=1}^M w^{k, t-1} K(\theta^{m,t}; \theta^{k, t-1})}
\end{equation*}
\State \textbf{output round $t$}: $\{ (\theta^{m,t}, w^{m, t}) \}_{m \in \llbracket 1, M \rrbracket}$
\EndFor
\State \textbf{return} accepted samples $\theta^n \sim p(\theta|\text{corr}(C^n,C_\text{ref}) \geq \epsilon_{\text{corr}})$
\end{algorithmic}
\end{algorithm}

When $\epsilon _{\text{corr}}\to 1$, $p(\theta|\text{corr}(C,C_\text{ref}) \geq \epsilon_{\text{corr}}) \to p(\theta | C_\text{ref})$.

\newpage
\subsection{With a summary statistic and the classical $l^2$-norm -- \ABCCNN \label{sec:ABCCNN_appendix}}
The other ABC approach uses a pre-learned summary statistic $S_\phi$.
\begin{algorithm}[H]
\caption{ABC with learned summary statistic inspired from \cite{summstatabc} \label{alg:abc-cnn}}
\begin{algorithmic}
\State \textbf{Input}:  (deep) neural network (DNN) $S_{\phi}$, threshold $\epsilon$, Euclidean norm in $\mathbb{R}^n$, simulator, prior $p$
\State \textbf{Return}:  Samples $\theta$ from the estimated posterior density $p(. \mid \Vert S_{\phi}(C)-S_{\phi}(C_\text{ref}) \Vert  \leq \epsilon)$\\ 
\State \textbf{Stage 1: learn the summary statistic} $S_{\phi}(.)$ s.t. $S_{\phi}(C) \approx \mathbb{E}\left[\theta|C\right]$
\State generate a train set $(\theta^n, C^n)$ from $p(\theta)p(C|\theta)$
\State train a DNN $S_\phi$ on this train set with the loss to minimize in $\phi$ \begin{equation*}
    \widehat{\mathcal{L}}_{\text{DNN}}(\phi)=\frac 1N \sum_{1\leq n\leq N}  \Vert S_{\phi}(C^n) - \theta^n \Vert ^2_ 2
\end{equation*}
\State output $S_{\phi}(.)$ s.t. $S_{\phi}(C) \approx \mathbb{E}\left[\theta|C\right]$
\State \textbf{Stage 2: run ABC with the learned summary statistic} $S_{\phi}$ \textbf{and the criterion} $ \Vert S_{\phi}(C)- S_{\phi}(C_\text{ref}) \Vert  \leq \epsilon$
\State \textbf{return} accepted samples $\theta^n \sim p(.  \mid \Vert S_{\phi}(C^n)-S_{\phi}(C_\text{ref}) \Vert  \leq \epsilon)$
\end{algorithmic}
\end{algorithm}
For $S_\phi$ informative enough, and when $\epsilon \to 0$, \begin{equation*}
    p(\theta  \mid \Vert S_{\phi}(C)-S_{\phi}(C_\text{ref}) \Vert  \leq \epsilon) \to p(\theta | S_{\phi}(C_\text{ref})) \approx p(\theta | C_\text{ref}).
\end{equation*}
\section{SNPE -- \SBICNN \label{sec:snpe_appendix}}
The last inference approach is SNPE based on normalizing flows and the pre-learned summary statistic $S_\phi$.\\
It is a sequential method: in the first round, $\theta$ is drawn from an uninformative prior. From the next rounds, $\theta$ is drawn from a proposal: the posterior density estimated from the previous round. This way, $\theta$ is more informative about $C_\text{ref}$ and the inference is expected to be refined across rounds.
\begin{algorithm}[H]
\caption{SNPE inspired from \cite{snpea} and \cite{snpec} \label{snpe}}
\begin{algorithmic}
\State \textbf{Input}:  $T$ rounds, posterior density estimator $p_{\psi}$, simulator, prior $p$, simulation budget $N$, observation $C_\text{ref}$, pre-learned summary statistic $S_\phi$
\State \textbf{Return}:  The estimated posterior density $p_\psi(.|S_\phi(C_\text{ref}))$\\ 
\For {round $t=1,...,T$}
\If{ $t=1$}
$p_t = p$
\EndIf
\For {$n=1,...,N$}
\State sample $\theta^n \sim p_t$
\State sample $C^n \sim p(.|\theta^n)$
\EndFor
\State train the posterior estimator $p_{\psi}$ on $\mathcal{D} = \{(\theta^n, C^n)\}_n$ with the loss to minimize in $\psi$ 
\begin{equation*}
    \widehat{\mathcal{L}}_{\text{NPE}}(\psi)=-\frac 1 N \sum_{1 \leq n \leq N} \log p_{\psi}(\theta^n|S_\phi(C^n))
\end{equation*}
\State use $p_{\psi}$ to construct the estimated posterior : $p_{\psi}(.|S_\phi(C_\text{ref}))$.
\State define the proposal for the next round : $p_t(\theta)  = p_{\psi}(\theta|S_\phi(C_\text{ref}))$
\EndFor
\State \textbf{return} samples $\theta^n \sim p_\psi(\theta|S_\phi(C_\text{ref}))$
\end{algorithmic}
\end{algorithm}

\section{Small genome inference \label{sec:3_chr_appendix}}
We work with the \textit{S. cerevisiae}'s first three chromosomes. $\theta$ is directly inferred from the entire contact map $C_\text{ref}$. We present a benchmark of metrics to assess the performance of the different inference methods : they evaluate both the proximity of the samples to $\theta_\text{ref}$ and the closeness of the densities to the `true' posterior $\delta_{\theta_\text{ref}}$.
\begin{figure}[H]
    \centering
    \includegraphics[width=12cm, height=9cm]{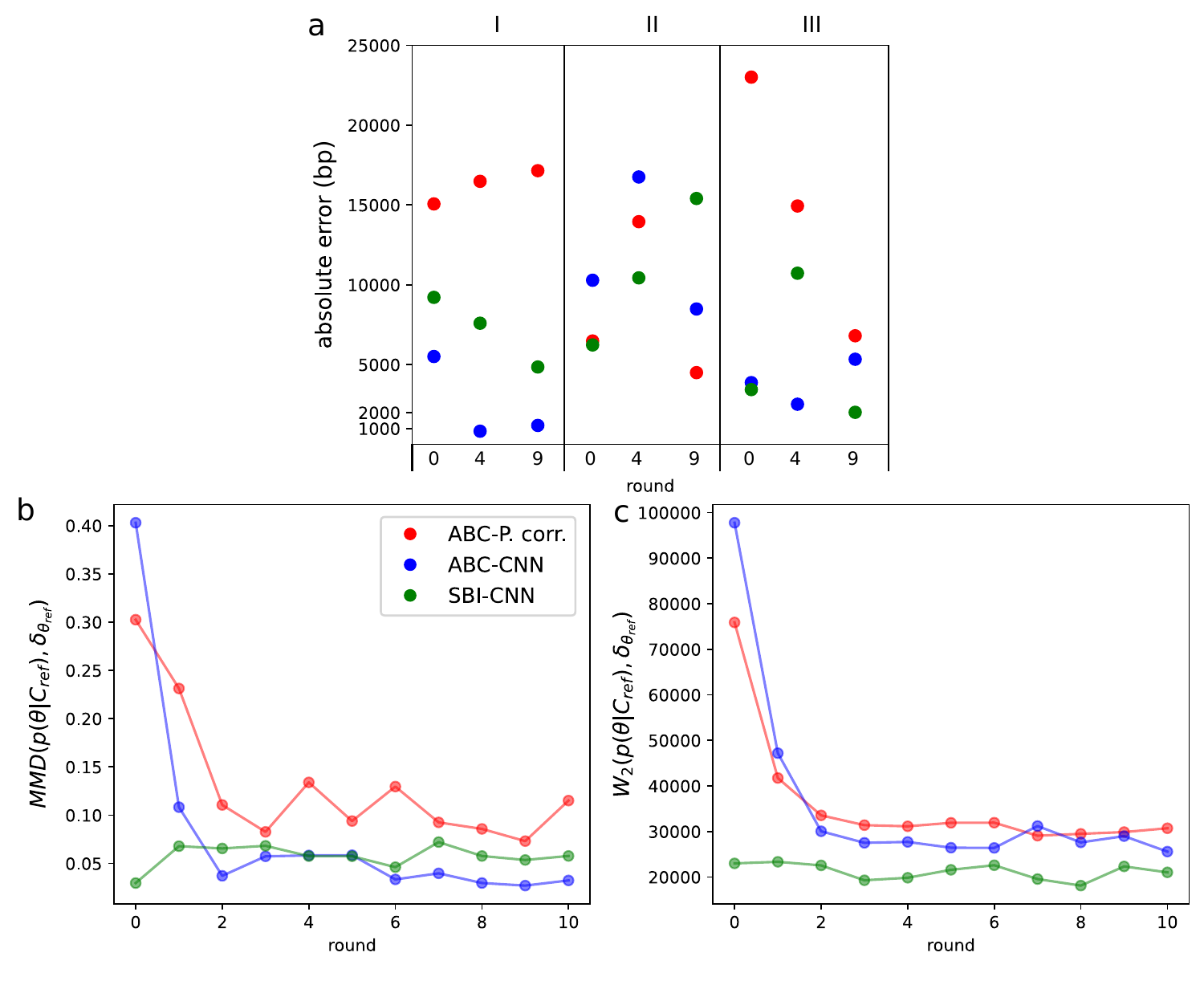}
    \caption{\label{fig:metrics_curves}\small We report the absolute error per dimension of $\theta$ between the mean computed over the $5\%$ best-performing samples and $\theta_\text{ref}$ (\textbf{a}) as well as the Maximum Mean Discrepancy (MMD) (\textbf{b}) and the Wasserstein-2 distance (\textbf{c}) between $p(\theta|C_\text{ref})$ and $\delta_{\theta_\text{ref}}$ \cite{mmd}.}
\end{figure}

\section{Whole genome inference \label{sec:16_chr_inference}}
To reduce the dimension of the problem, we carry 16 parallel inferences: one per dimension of $\theta$. Thus, we have 16 $1D$ inference problems where the parameter $\theta_i$ is drawn from a Uniform prior whose range is the size of the chromosome $i$ in bp. The simulator creates the $i^\mathrm{th}$ row of trans-contact blocks of a contact map $C$ (denoted $C_i)$. All the inference methods target the posterior $p(\theta_i | C_{\text{ref},i})$. We need also to learn 16 summary statistics $\{S_{\phi_i}\}_i$ to project each row of trans-contact blocks $C_i$ to $\theta_i$.\\
$S_{\phi_i}$ is a CNN to capture the information of $C_i$ followed by an MLP to project this information into $\theta_i$. On the one hand, as the rows of trans-contact blocks $C_i$ are quite similar, we choose a shared architecture for the CNN between chromosomes. On the other hand, each MLP depends on the size of each chromosome so a chromosome-specific architecture is thus needed for this part of the DNN.\\
For the SBI method, we also need to learn $16$ normalizing flows. As for the 3-chromosomes case, we choose a MAF as well as SNPE-C for the experiments. 

\begin{figure}[H]
    \centering
    \includegraphics[width=14cm, height=8cm]{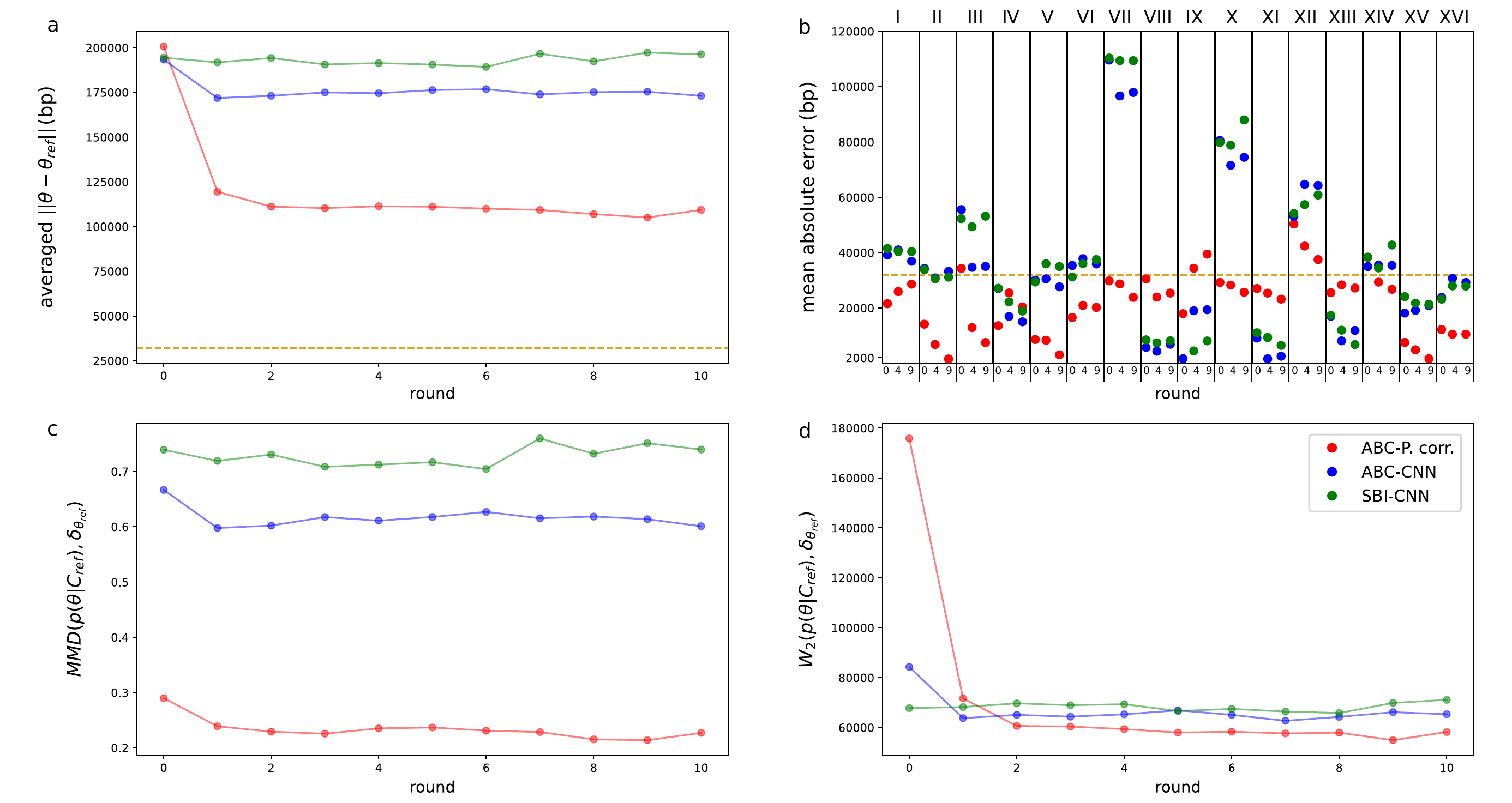}
    \caption{\label{fig:metrics_16_chr}\small We report the mean Euclidean distance between $\theta$ and $\theta_\text{ref}$ (\textbf{a}), computed over the $5\%$ best-performing samples, the absolute error per dimension of $\theta$ between the mean $\theta$ computed over the $5\%$ best-performing samples and $\theta_\text{ref}$ (\textbf{b})  as well as the MMD (\textbf{c}) and the Wasserstein-2 distance (\textbf{d}) between $p(\theta|C_\text{ref})$ and $\delta_{\theta_\text{ref}}$. The horizontal dotted line stands for the resolution of the contact map $C_\text{ref}$ (in bp) in the top figures.}
\end{figure}

\end{document}